\documentclass[a4paper]{article}
\usepackage{float}
\usepackage{xcolor}
\usepackage{soul}
\usepackage{dblfloatfix}
\usepackage{INTERSPEECH2022}
\usepackage{url}

\title{Shallow Fusion of Weighted Finite-State Transducer and Language Model for Text Normalization}
\name{Evelina Bakhturina*\thanks{*Equal contribution. Order determined alphabetically.\\ Preprint. Submitted to INTERSPEECH-22}, Yang Zhang*\footnotemark[1], Boris Ginsburg}
\address{
  NVIDIA, Santa Clara, USA}
\email{\{ebakhturina, yangzhang, bginsburg\}@nvidia.com}

\begin{document}

\maketitle

\begin{abstract}

Text normalization (TN) systems in production are largely rule-based using weighted finite-state transducers (WFST). However, WFST-based systems struggle with ambiguous input when the normalized form is context-dependent.  On the other hand, neural text normalization systems can take context into account but they suffer from unrecoverable errors and require labeled normalization datasets, which are hard to collect. We propose a new hybrid approach that combines the benefits of rule-based and neural systems. First, a non-deterministic WFST outputs all normalization candidates, and then a neural language model picks the best one -- similar to shallow fusion for automatic speech recognition. While the WFST prevents unrecoverable errors, the language model resolves contextual ambiguity. The approach is easy to extend and we show it is effective. It achieves comparable or better results than existing state-of-the-art TN models.

\end{abstract}


\section{Introduction}

\begin{figure}[H]
  \centering
  \includegraphics[width=\linewidth]{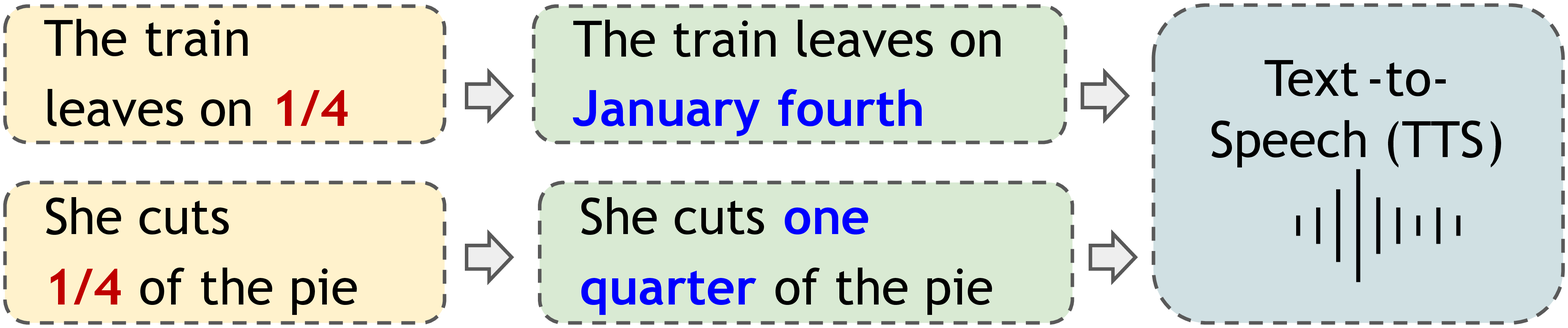}
  \caption{The normalization of the semiotic token  ``1/4" is context-dependent.}
  \label{fig:context_normalization_examples}
\end{figure}

Text normalization (TN) converts text from the canonical written to the spoken form, and it is an essential pre-processing step of text-to-speech (TTS) systems, see Figure~\ref{fig:context_normalization_examples}.
Sentences consist of plain and special words or phrases, also known as \textit{semiotic tokens}~\cite{taylor_tts}. Semiotic tokens can be grouped into classes like cardinals, fractions, or dates. Semiotic tokens require non-trivial normalization while plain words remain the same. In the example shown in Figure~\ref{fig:context_normalization_examples} \textit{``1/4"} is the only semiotic token and thus needs to be normalized.

    

TN has a few characteristics that differentiate it from other natural language processing (NLP) tasks. First, the normalized texts are scarce and difficult to collect~\cite{5700892, ebden2015kestrel}. To our knowledge, Google text normalization dataset~\cite{Sproat2017} is the only sizeable public dataset. Second, TN applications such as mobile assistants demand very low error rates~\cite{zhang2019neural}. Errors are called \textit{unrecoverable} if they alter the input semantic~\cite{zhang2019neural}, e.g. \(  \textit{ ``1/4"}  \rightarrow\ \textit{ ``one fourteenth"}\). Information-sensitive industry sectors such as finance can not tolerate any chance of unrecoverable errors. Lastly, semiotic tokens make up only a small percentage of the sentence, but they stem from a huge variety of semiotic classes~\cite{zhang2019neural}. These make TN a challenging NLP task, and it remains an active area of research.


\begin{figure*}[!tb]
  \centering
  \includegraphics[width=0.9\textwidth]{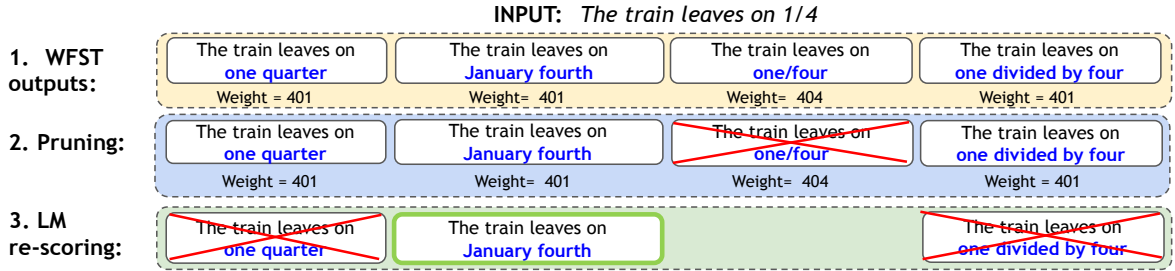}
  \caption{WFST+LM shallow fusion pipeline. 1. WFST generates all possible normalization forms and assigns weights to each option. 2. Pruning normalization options with weights above threshold ``401.2", in this example we drop  ``one/four". It has higher weight since it was not fully normalized. 3. LM re-scoring picks the best among the remaining options.}
  \label{fig:approach_overview}
\end{figure*}

\begin{figure*}[!tb]
  \centering
  \includegraphics[width=\linewidth]{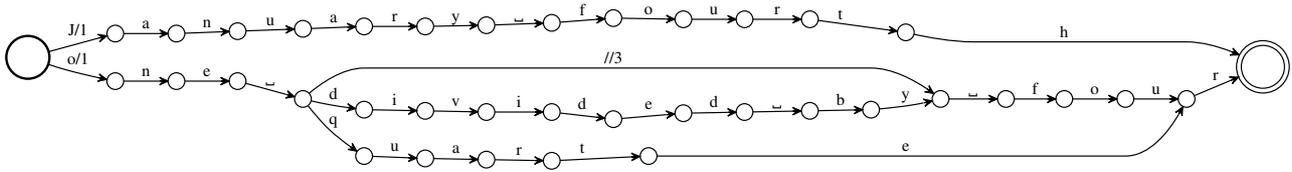}
  \caption{WFST lattice of normalization options for  ``1/4". The options shown are ``January fourth"(W=1),  ``one quarter"(W=1), ``one divided by four"(W=1),  ``one/four"(W=4). By construction, if a semiotic token is not fully normalized, e.g.  ``/" in  ``one/four", it has a higher weight, e.g. 4 vs. 1. This allows us to filter out these options by applying a threshold on the weights.}
  \label{fig:fst}
\end{figure*}

TN systems can be rule-based, neural or a combination of both. The traditional \textbf{rule-based} systems utilize either regular expressions or weighted finite-state transducers (WFST)~\cite{Mohri2009} to define a set of language-specific rules. These systems are easy to extend, debug and reason about. However, they can not resolve contextual ambiguity well without making the rules significantly more complex. For example, it is very complicated to write a rule that resolves \textit{ ``1/4"} to \textit{``one quarter"} in one case and to \textit{``January fourth"} in another context, see Figure \ref{fig:context_normalization_examples}. Additionally, rule-based systems can be easily customized to support Speech Synthesis Markup Language (SSML)~\cite{ssml}, which is becoming a requirement for TTS systems in production. SSML gives users control over different verbalization options, and users can explicitly specify if \textit{``2"} should be spoken as the cardinal \textit{``two"} or the ordinal \textit{``second"}.
Recent \textbf{neural} systems treat TN as a machine translation task using Seq2Seq models~\cite{Sproat2017, mansfield-etal-2019-neural, lai2021unified}. These are good at incorporating context, but they require large datasets to scale, 
and suffer from catastrophic unrecoverable errors. Therefore, some \textbf{hybrid} systems propose to use a generative neural network (NN) constrained by WFST grammars~\cite{zhang2019neural, sunkara2021neural}. Nonetheless, if the WFST does not cover a particular example, the NN performs unconstrained normalization. These hybrid systems can reduce the number of unrecoverable errors, but still require supervised training data~\cite{zhang2019neural}. Therefore, state-of-the-art TN systems in production rely mostly on rule-based normalization~\cite{ebden2015kestrel, Mohri2009}.

We propose a new hybrid TN approach that uses rule-based WFST grammars to output all possible normalization options, which are then re-scored with a pre-trained neural language model (LM) to resolve contextual ambiguities, see Figure \ref{fig:approach_overview}. This approach has the following benefits:

\begin{itemize}
    \item Ability to control system's output and prevent certain unrecoverable errors, like hallucinations.
    \item Easy to improve normalization capability by extending the WFST grammar.
    \item Compatible with many off-the-shelf pre-trained LMs. Also, LMs can be easily adapted as they are trained in a self-supervised fashion without the need for labeled data, i.e. written-spoken text pairs.
\end{itemize}


The proposed WFST+LM approach is similar to LM-decoding in automatic speech recognition (ASR) or machine translation~\cite{shallowfusion}. Instead of a single transcription, ASR outputs multiple options, and then a LM model re-scores these options to account for language fluency and context. To the best of our knowledge, we are the first to apply shallow fusion of WFST and LM to TN. We created a hand-crafted set of context-dependent normalization test cases to demonstrate the effectiveness of our approach compared to rule-based and NN-based TN systems. The model and the set of the context-dependent test cases will be released at \url{https://github.com/NVIDIA/NeMo}. 


\section{WFST-based TN re-scoring with LM}
\label{wfst_overview}


\subsection{Deterministic WFST}
A path through a weighted finite-state transducer (WFST)~\cite{Mohri2009} maps an input sequence to an output sequence, where each transduction, called arcs, has an associated weight \textbf{w}, see Figure~\ref{fig:fst}. A path's weight \textbf{W} is the sum of all its arcs' weights, and a deterministic WFST outputs a single shortest path which has the minimum weight.  
As a starting point for our proposed system we use a deterministic WFST based on the NeMo text normalization framework~\cite{zhang2021nemo}, which does not take context into account (hereafter DetWFST). In case of ambiguous inputs such as \textit{``1/4"} in \textit{``The trains leaves on 1/4"}, a deterministic WFST can mistakenly choose \textit{``one quarter"} over \textit{``January fourth"}.

\subsection{Non-deterministic WFST}
We extend aforementioned DetWFST to output multiple valid normalization options, see Figure~\ref{fig:approach_overview}. We eliminate redundancy of semantically equivalent normalization forms, e.g., \textit{``January fourth"} and \textit{``the fourth of January"}. 
Beyond that, we want every semiotic token normalized, so WFST assigns higher weights to a path that keeps the token unchanged. Specifically, we picked a weight of \textbf{w}=100 for unchanged word and \textbf{w}=2 for a punctuation mark. On the other hand, all arcs that normalize a semiotic token have weights \textbf{w}$ \in $[1.0; 1.01].
This allows us to prune all unwanted outputs by enforcing a threshold on their respective paths' weights, see Figures~\ref{fig:approach_overview},~\ref{fig:fst}.

Let us assume our input has $k$ semiotic tokens, and $S$ is the set of possible normalization options, where ${w_{s,k}}$ is the weight of the $k$-th token of the normalization option $s$. Then, all weights $W_{s}$ for $s \in S$ satisfy the following condition:

\begin{align*}
    W_{s} = C + \sum_k{w_{s,k}}
     &\leq C + k \cdot 1.01 \\
     &= C + k \cdot 1.0 + k \cdot 0.01
\end{align*}
where $C$ is the total weight of all non-semiotic tokens, and we assume that $k$ is at most $20$. Further,

\begin{align*}
    C + k \cdot 1.0 + k \cdot 0.01
     &\leq \min_{s \in S} W_s + k \cdot 0.01 \\
     &\leq \min_{s \in S} W_s + 20 \cdot 0.01 \\
     &= \min_{s \in S} W_s + 0.2 
\end{align*}
Thus, we are keeping only outputs with weights within $0.2$ of the shortest path.

\subsection{Language model re-scoring}

The pruned WFST output options are re-scored with pre-trained language models from the Hugging Face library~\cite{wolf-etal-2020-transformers}.
For autoregressive language models, like GPT~\cite{radford2018improving}, we rank the normalization options based on their perplexity. For masked language models we use Masked Language Model (MLM) scoring\footnote{\url{https://github.com/awslabs/mlm-scoring}}~\cite{salazar2019masked}. MLM is the average of pseudo-log-likelihood scores calculated by masking input tokens one by one. 

We modify MLM score calculation when an input sentence contains multiple semiotic tokens that require normalization. In this case, to minimize bias, we exclude all but one semiotic token from the context, which LM considers. The goal is to have only a single semiotic option when calculating the MLM score, while the rest of the semiotic tokens in question are masked. The scores for each masked variant are averaged to get an aggregated value for the sentence. For example, to calculate the score for \textit{``What's $<$ one half $>$ cup plus $<$ two thirds $>$ cup?"}, where the angular brackets highlight semiotic spans, we first calculate weights for \textit{``What's [MASK] cup plus two thirds cup?"} and \textit{``What's one half cup plus [MASK] cup?"}. Then, we take the average to get the score for the sentence of interest.

\section{Experiment setup}
\label{experiments}

\subsection{Evaluation datasets}
\label{datasets}

We use pairs of written and normalized English texts for evaluation. 
Publicly available ``general" datasets are large and suit as a benchmark to compare our results with related works. We additionally created a ``confusing" dataset, which focuses on ambiguous examples.

Google text normalization dataset (hereafter GoogleTN)~\cite{Sproat2017} is one of the few publicly available TN datasets for English with decent coverage of various semiotic classes and is widely used in the field.
Normalization forms in this dataset are generated semi-automatically with Google
TTS system’s Kestrel text normalization system~\cite{ebden2015kestrel}. The test set contains 7551 sentences. For our second ``general" dataset, we use LibriTTS~\cite{zen2019libritts} derived from LibriVox audiobooks\footnote{\url{https://librivox.org/}} and corresponding Gutenberg texts\footnote{\url{ https://www.gutenberg.org/}}. We only evaluate the subset of LibriTTS where the normalization differs from the written form, resulting in 7677 sentences.

For the ``confusing" dataset (hereafter EngConf) we collected context-dependent normalization examples from an internal dataset, Google TN dataset (disjoint from aforementioned) and a conversational corpus~\cite{XLiu.etal:IWSDS2019}. Since the Google TN data was labeled semi-automatically there are many errors. These were manually corrected before being added to EngConf.

\subsection{Models}

We compare our WFST+LM approach with the deterministic WFST (DetWFST)~\cite{zhang2021nemo} and the end-to-end neural TN model (Duplex)~\cite{lai2021unified}. Both baselines achieve one of the the highest reported normalization accuracies on GoogleTN dataset~\cite{Sproat2017}. Duplex comprises a distilled RoBERTa-based token classification model~\cite{liu2019roberta} and a generative T5-base model~\cite{raffel2019exploring}. Every input token the classifier identifies as semiotic token is normalized by the T5 model.
In our hybrid approach we used the pre-trained BERT-based-uncased~\cite{devlin2018bert} checkpoint from Hugging face~\cite{wolf-etal-2020-transformers}, unless stated otherwise.

\subsection{Metric}

For quantitative analysis we calculate the sentence-level accuracy from the ground truth normalization and the model output~\cite{Sproat2017}. We relax the strict string equality requirement by accepting different formats of the same phrase as equivalent. For example, \textit{``the fifth of November two thousand twenty"} and \textit{``November fifth twenty twenty"} are both valid spoken forms of the date \textit{``2020/11/05"} and are considered equivalent.

\section{Results} \label{sec:results}


\subsection{Overall performance}

Table \ref{tab:results} summarizes the performance results of WFST+LM, DetWFST and Duplex on the datasets EngConf, GoogleTN, and LibriTTS. For WFST+LM, we found masked language models with MLM-scoring~\cite{salazar2019masked} work better for than autoregressive models, e.g. GPT~\cite{radford2018improving} and DistilGPT2~\cite{distilgpt2}, see Table~\ref{tab:models}. BERT-base-uncased~\cite{devlin2018bert} results in the highest sentence accuracy.

\begin{table}[t]
\centering
\caption{Comparison of the model performance of DetWFST, Duplex and WFST+LM on three datasets (sentence accuracy, \%). The EngConf contains confusing ``one-to-many" test cases, the other datasets cover general TN cases.}
\label{tab:results}
    \begin{tabular}{lcccc}
    \toprule
\textbf{Dataset}  & \begin{tabular}[c]{@{}l@{}}\textbf{Number of}\\ \textbf{sentences}\end{tabular} & \begin{tabular}[c]{@{}c@{}}\textbf{Det}\\ \textbf{WFST}\end{tabular} & \textbf{Duplex}  & \begin{tabular}[c]{@{}l@{}}\textbf{WFST} \\ + \textbf{LM}\end{tabular} \\\toprule
EngConf  & 231          & 68.83       & 55.41    & \textbf{94.37}    \\
GoogleTN & 7551         & 97.29       & \textbf{99.07}     & 97.79    \\
LibriTTS & 7677         & 98.65       & 90.40     & \textbf{99.01}    \\
\bottomrule
\end{tabular}
\end{table}

\begin{table}[t]
\centering
\caption{Sentence accuracies on EngConf dataset using different language models for LM re-scoring.}
\label{tab:models}
\begin{tabular}{lc}
\toprule
\textbf{Model} & \textbf{accuracy, \%}  \\
\toprule
distilgpt2~\cite{distilgpt2}                  &               82.68                     \\
openai-gpt~\cite{openai-gpt}                  &               88.31               \\
bert-base-en-uncased (MLM)~\cite{devlin2018bert}  &               \textbf{94.37}                    \\
bert-base-en-cased (MLM)~\cite{devlin2018bert}    &               92.64                    \\
roberta-base-en-cased (MLM)~\cite{liu2019roberta} &              93.07                \\ \bottomrule                     
\end{tabular}
\end{table}

\begin{table*}[]
\centering
\caption{Error patterns for DetWFST, Duplex and WFST+LM.}
\label{tab:errors}
\definecolor{myreen}{RGB}{50,205,50}
\definecolor{myorange}{RGB}{255,165,0}
\newcommand{\notaff}{\color{myreen}\begin{tabular}[c]{@{}c@{}}not\\affected\end{tabular}}
\newcommand{\lessaff}{\color{myorange}\begin{tabular}[c]{@{}c@{}}less\\affected\end{tabular}}
\newcommand{\aff}{\color{red}affected}
\begin{tabular}{lcccc}
\toprule
\textbf{Error Type} & \textbf{\begin{tabular}[c]{@{}c@{}}Det\\WFST\end{tabular}} & \textbf{Duplex} & \textbf{\begin{tabular}[c]{@{}c@{}}WFST\\+LM\end{tabular}} & \textbf{Example errors}  \\
\toprule
Number error  & \notaff & \aff & \notaff & 
\begin{tabular}{p{15mm}p{70mm}@{}ll@{}}
IN:&    Number 10001 \\ 
TARGET:&    Number Ten thousand one \\ 
Duplex:&    Number {\color{red}one hundred} thousand one\end{tabular} \\

\hline
Unknown format  & \aff & \aff & \aff & 
\begin{tabular}{p{15mm}p{70mm}@{}ll@{}}
IN:& Set the thermostat to 75F \\ 
TARGET:& Set the thermostat to seventy five degrees Fahrenheit\\
DetWFST:& Set the thermostat to seventy five {\color{red}F} \\
Duplex:& Set the thermostat to seventy five {\color{red}degrees}\\
WFST+LM:& Set the thermostat to seventy five {\color{red}F}
\end{tabular} \\

\hline
Hallucination  & \notaff & \aff & \notaff & 
\begin{tabular}{p{15mm}p{70mm}@{}ll@{}}


IN: & Josiah in the gutter! exclaimed Mrs. Pegler. \\
TARGET: & Josiah in the gutter! exclaimed Misses Pegler. \\
Duplex: & Josiah in the gutter! exclaimed {\color{red} m r e} Pegler. \end{tabular} \\

\hline
Omission  & \notaff & \aff & \notaff & 
\begin{tabular}{p{15mm}p{70mm}@{}ll@{}}
IN:& Cambridgeshire, CB 10 1 SD\\ 
TARGET:& Cambridgeshire c b one zero one s d \\
Duplex:& Cambridgeshire c b {\color{red}one one} s d \\
\end{tabular} \\

\hline
Class ambiguity  & \aff & \lessaff & \lessaff & 
\begin{tabular}{p{15mm}p{70mm}@{}ll@{}}
IN:& The train leaves on 1/4\\ 
TARGET:& The train leaves on January fourth\\
DetWFST:& The train leaves on {\color{red}one quarter} \\
Duplex:& The train leaves on {\color{red}one quarter}  \\

\end{tabular} \\


\hline
Smart URL splitting  & \aff & \lessaff & \lessaff & 
\begin{tabular}{p{15mm}p{70mm}@{}ll@{}}
IN:& It's WeAreSC.com\\ 
TARGET:& It's We Are S C dot com\\
DetWFST:& It's {\color{red} W e A r e} s c dot com \\
Duplex:& It's We Are {\color{red} esc} dot com \\
WFST+LM: & It's we are {\color{red}sc} dot com \\
\end{tabular} \\

\bottomrule
\end{tabular}
\end{table*}

The WFST+LM beats Duplex model on all but GoogleTN data. Duplex, which is a neural model trained on 800,000 sentences of the training split of the Google text normalization dataset, reaches an accuracy of 99.07\% compared to 97.29\% of DetWFST and 97.79\% of WSFT+LM. To our knowledge, the accuracy of DetWFST is the highest among publicly available rule-based systems on GoogleTN and is only slightly behind the neural model. Both DetWFST and WFST+LM lack some rules for exceptional input formats, which account for some of the performance gap, e.g., \textit{``Set the thermostat to 75F"} in Table~\ref{tab:errors}. 
However, Duplex fails to generalize on LibriTTS and EngConf datasets. WFST+LM outperforms Duplex with 99.01\% vs. 90.40\% and 94.37\% vs. 55.41\% sentence accuracy, respectively. Duplex suffers from unknown input formats and domain shifts. For example, it does not recognize \textit{``1/4"} as a date in \textit{``The train leaves on 1/4"} and falsely produces \textit{``The train leaves on one quarter"} instead. Duplex can also make mistakes when normalizing cardinals containing more than four digits, e.g. \textit{``10001"} $\rightarrow$ \textit{``one hundred thousand one"}. For information-sensitive industries such as finance, this type of unrecoverable error is detrimental. More failure types can be found in Table \ref{tab:errors}.

WFST+LM outperforms DetWFST on all three datasets. On the general domain datasets, LibriTTS and GoogleTN, the LM re-scoring does not provide much performance gain over the deterministic WFST. These datasets do not contain many confusing cases and thus can be mostly normalized using deterministic grammars. However, LM re-scoring offers a noticeable improvement over DetWFST on the EngConf dataset. Language model integration resolves ambiguities from context and achieves 94.37\% vs. 68.83\% sentence accuracy. One such context-dependent example is \textit{``The train leaves on 1/4"} shown in Figure~\ref{fig:approach_overview}.

Similar to shallow fusion~\cite{shallowfusion} for speech recognition or neural machine translation, LM re-scoring can make mistakes especially if the LM was trained on a different domain. In our case, publicly available pre-trained LMs are trained on un-normalized data but used to score normalized text. For example, the LM struggles to differentiate between \textit{``born in one thousand nine hundred seventy"} vs. \textit{``born in nineteen seventy"} as a spoken form of \textit{``born in 1970"} and \textit{``Henry III"} can be mistakenly normalized to \textit{``Henry three"} instead of \textit{``Henry the third"}. However, while the failures of the re-scoring approach can be expected by design of the WFST output options, it is not the case for the neural model.  
Failing examples for each of the methods are listed in Table~\ref{tab:errors}.

Neural models such as Duplex can make unexpected errors and are hard to control and debug. At the same time, the WFST-based solutions can fail to cover some cases due to a limited set of grammar rules, but they are reliable and easy to debug if a rule is present.


\subsection{Additional use cases of re-scoring}

LM re-scoring can help choose the right inflection form for morphology rich languages. For example, in German, ordinals change their suffix depending on the following noun.  

Email and URL normalization can also benefit from the proposed approach. Without an extra magnitude of complexity, deterministic WFSTs can only convert URLs into space-separated characters. For example, \textit{``yemail@greattech.com"} becomes \textit{``m y e m a i l at g r e a t t e c h dot com"}. However, URLs often contain words which should be pronounced as a whole, e.g. \textit{``my email at great tech dot com"}. LM re-scoring can up-vote this option easily. Duplex can sometimes identify words in a URL correctly but it is prone to hallucinations or omissions especially when encountering unseen words, see Table~\ref{tab:errors}.


\section{Conclusion}

We propose shallow fusion of WFST and neural language model for text normalization. The non-deterministic WFST outputs multiple normalization options for ambiguous inputs, which are re-scored by a pre-trained language model. WFST+LM provides contextual disambiguation while preventing unrecoverable errors such as hallucination and token omission that are common for neural text normalization approaches. The WFST+LM system is easy to debug and extend, and it allows users to remain in control. We find that masked language models outperform generative models in our experiments. 
This approach reaches an accuracy of 94.37\% on an ambiguous context-dependent dataset, which outperforms a state-of-the-art neural model, and it reaches a competitive 97.79\% on the Google text normalization dataset. WFST+LM achieves 99.01\% accuracy on the LibriTTS corpus, showing its resilience to domain shift.



\section{Acknowledgements}

The authors would like to thank Vitaly Lavrukhin and Elena Rastorgueva for their review and feedback.

\pagebreak
\bibliographystyle{IEEEtran}
\bibliography{template}

\begin{thebibliography}{10}
\providecommand{\url}[1]{#1}
\csname url@samestyle\endcsname
\providecommand{\newblock}{\relax}
\providecommand{\bibinfo}[2]{#2}
\providecommand{\BIBentrySTDinterwordspacing}{\spaceskip=0pt\relax}
\providecommand{\BIBentryALTinterwordstretchfactor}{4}
\providecommand{\BIBentryALTinterwordspacing}{\spaceskip=\fontdimen2\font plus
\BIBentryALTinterwordstretchfactor\fontdimen3\font minus
  \fontdimen4\font\relax}
\providecommand{\BIBforeignlanguage}[2]{{%
\expandafter\ifx\csname l@#1\endcsname\relax
\typeout{** WARNING: IEEEtran.bst: No hyphenation pattern has been}%
\typeout{** loaded for the language `#1'. Using the pattern for}%
\typeout{** the default language instead.}%
\else
\language=\csname l@#1\endcsname
\fi
#2}}
\providecommand{\BIBdecl}{\relax}
\BIBdecl

\bibitem{taylor_tts}
P.~Taylor, \emph{Text-to-Speech Synthesis}.\hskip 1em plus 0.5em minus
  0.4em\relax Cambridge University Press, 2009.

\bibitem{5700892}
R.~{Sproat}, ``Lightly supervised learning of text normalization: Russian
  number names,'' in \emph{IEEE Spoken Language Technology Workshop}, 2010.

\bibitem{ebden2015kestrel}
P.~Ebden and R.~Sproat, ``The kestrel tts text normalization system,''
  \emph{Natural Language Engineering}, 2015.

\bibitem{Sproat2017}
R.~Sproat and N.~Jaitly, ``An {RNN} model of text normalization,'' in
  \emph{Interspeech}, 2017.

\bibitem{zhang2019neural}
H.~Zhang, R.~Sproat, A.~H. Ng, F.~Stahlberg, X.~Peng, K.~Gorman, and B.~Roark,
  ``Neural models of text normalization for speech applications,''
  \emph{Computational Linguistics}, 2019.

\bibitem{Mohri2009}
M.~Mohri, \emph{Weighted Automata Algorithms}, 2009, pp. 213--254.

\bibitem{ssml}
``{Speech Synthesis Markup Language (SSML)},''
  \url{https://www.w3.org/TR/speech-synthesis}.

\bibitem{mansfield-etal-2019-neural}
C.~Mansfield, M.~Sun, Y.~Liu, A.~Gandhe, and B.~Hoffmeister, ``Neural text
  normalization with subword units,'' in \emph{Proceedings of the 2019
  Conference of the North {A}merican Chapter of the Association for
  Computational Linguistics: Human Language Technologies, Volume 2 (Industry
  Papers)}, 2019.

\bibitem{lai2021unified}
T.~M. Lai, Y.~Zhang, E.~Bakhturina, B.~Ginsburg, and H.~Ji, ``A unified
  transformer-based framework for duplex text normalization,'' \emph{arXiv
  preprint arXiv:2108.09889}, 2021.

\bibitem{sunkara2021neural}
M.~Sunkara, C.~Shivade, S.~Bodapati, and K.~Kirchhoff, ``Neural inverse text
  normalization,'' in \emph{IEEE International Conference on Acoustics, Speech
  and Signal Processing (ICASSP)}, 2021.

\bibitem{shallowfusion}
{\c{C}}.~G{\"{u}}l{\c{c}}ehre, O.~Firat, K.~Xu, K.~Cho, L.~Barrault, H.~Lin,
  F.~Bougares, H.~Schwenk, and Y.~Bengio, ``On using monolingual corpora in
  neural machine translation,'' \emph{CoRR}, vol. abs/1503.03535, 2015.

\bibitem{zhang2021nemo}
Y.~Zhang, E.~Bakhturina, and B.~Ginsburg, ``{NeMo (Inverse) Text Normalization:
  From Development to Production},'' in \emph{Interspeech}, 2021.

\bibitem{wolf-etal-2020-transformers}
T.~Wolf, L.~Debut, V.~Sanh, J.~Chaumond, C.~Delangue, A.~Moi, P.~Cistac,
  T.~Rault, R.~Louf, M.~Funtowicz, J.~Davison, S.~Shleifer, P.~von Platen,
  C.~Ma, Y.~Jernite, J.~Plu, C.~Xu, T.~Le~Scao, S.~Gugger, M.~Drame, Q.~Lhoest,
  and A.~Rush, ``Transformers: State-of-the-art natural language processing,''
  in \emph{Proceedings of the 2020 Conference on Empirical Methods in Natural
  Language Processing: System Demonstrations}.\hskip 1em plus 0.5em minus
  0.4em\relax Association for Computational Linguistics, Oct. 2020, pp. 38--45.

\bibitem{radford2018improving}
A.~Radford, K.~Narasimhan, T.~Salimans, and I.~Sutskever, ``Improving language
  understanding by generative pre-training,'' 2018.

\bibitem{salazar2019masked}
J.~Salazar, D.~Liang, T.~Q. Nguyen, and K.~Kirchhoff, ``Masked language model
  scoring,'' in \emph{Proceedings of the 58th Annual Meeting of the Association
  for Computational Linguistics}, 2020.

\bibitem{zen2019libritts}
H.~Zen, V.~Dang, R.~Clark, Y.~Zhang, R.~J. Weiss, Y.~Jia, Z.~Chen, and Y.~Wu,
  ``Libritts: A corpus derived from librispeech for text-to-speech,'' in
  \emph{Interspeech}, 2019.

\bibitem{XLiu.etal:IWSDS2019}
X.~Liu, A.~Eshghi, P.~Swietojanski, and V.~Rieser, ``Benchmarking natural
  language understanding services for building conversational agents,'' in
  \emph{IWSDS}, 2019.

\bibitem{liu2019roberta}
Y.~Liu, M.~Ott, N.~Goyal, J.~Du, M.~Joshi, D.~Chen, O.~Levy, M.~Lewis,
  L.~Zettlemoyer, and V.~Stoyanov, ``Roberta: A robustly optimized bert
  pretraining approach,'' \emph{arXiv preprint arXiv:1907.11692}, 2019.

\bibitem{raffel2019exploring}
C.~Raffel, N.~Shazeer, A.~Roberts, K.~Lee, S.~Narang, M.~Matena, Y.~Zhou,
  W.~Li, and P.~J. Liu, ``Exploring the limits of transfer learning with a
  unified text-to-text transformer,'' \emph{Journal of Machine Learning
  Research}, 2020.

\bibitem{devlin2018bert}
J.~Devlin, M.-W. Chang, K.~Lee, and K.~Toutanova, ``{BERT}: Pre-training of
  deep bidirectional transformers for language understanding,'' in
  \emph{Proceedings of the 2019 Conference of the North {A}merican Chapter of
  the Association for Computational Linguistics: Human Language Technologies},
  2019.

\bibitem{distilgpt2}
``{DistilGPT2},'' \url{https://huggingface.co/distilgpt2}.

\bibitem{openai-gpt}
``{OpenAI GPT},''
  \url{https://huggingface.co/docs/transformers/model_doc/openai-gpt}.

\end{thebibliography}


\end{document}